\documentclass[letterpaper]{article} % DO NOT CHANGE THIS
\usepackage{aaai2026}  % DO NOT CHANGE THIS
\usepackage{times}  % DO NOT CHANGE THIS
\usepackage{helvet}  % DO NOT CHANGE THIS
\usepackage{courier}  % DO NOT CHANGE THIS
\usepackage[hyphens]{url}  % DO NOT CHANGE THIS
\usepackage{graphicx} % DO NOT CHANGE THIS
\usepackage{natbib}  % DO NOT CHANGE THIS AND DO NOT ADD ANY OPTIONS TO IT
\usepackage{caption} % DO NOT CHANGE THIS AND DO NOT ADD ANY OPTIONS TO IT
\usepackage{algorithm}
\usepackage{algorithmic}

\usepackage{newfloat}
\usepackage{listings}
\DeclareCaptionStyle{ruled}{labelfont=normalfont,labelsep=colon,strut=off} % DO NOT CHANGE THIS
\floatstyle{ruled}
\newfloat{listing}{tb}{lst}{}
\floatname{listing}{Listing}
\usepackage{booktabs}
\usepackage{amsmath}
\usepackage{amsfonts}
\usepackage{multirow}
\usepackage{svg}

\title{Graph VQ-Transformer (GVT): Fast and Accurate Molecular Generation via High-Fidelity Discrete Latents}
\author{
    Haozhuo Zheng,
    Cheng Wang,
    Yang Liu\thanks{Corresponding author.}
}
\affiliations{
    Harbin Institute of Technology\\
    24B903065@stu.hit.edu.cn, 21S003047@stu.hit.edu.cn, liuyang@hit.edu.cn
}

\begin{document}

\maketitle

\begin{abstract}
The de novo generation of molecules with desirable properties is a critical challenge, where diffusion models are computationally intensive and autoregressive models struggle with error propagation. In this work, we introduce the Graph VQ-Transformer (GVT), a two-stage generative framework that achieves both high accuracy and efficiency. The core of our approach is a novel Graph Vector Quantized Variational Autoencoder (VQ-VAE) that compresses molecular graphs into high-fidelity discrete latent sequences. By synergistically combining a Graph Transformer with canonical Reverse Cuthill-McKee (RCM) node ordering and Rotary Positional Embeddings (RoPE), our VQ-VAE achieves near-perfect reconstruction rates. An autoregressive Transformer is then trained on these discrete latents, effectively converting graph generation into a well-structured sequence modeling problem. Crucially, this mapping of complex graphs to high-fidelity discrete sequences bridges molecular design with the powerful paradigm of large-scale sequence modeling, unlocking potential synergies with Large Language Models (LLMs). Extensive experiments show that GVT achieves state-of-the-art or highly competitive performance across major benchmarks like ZINC250k, MOSES, and GuacaMol, and notably outperforms leading diffusion models on key distribution similarity metrics such as FCD and KL Divergence. With its superior performance, efficiency, and architectural novelty, GVT not only presents a compelling alternative to diffusion models but also establishes a strong new baseline for the field, paving the way for future research in discrete latent-space molecular generation.
\end{abstract}

\begin{links}
    % TODO: Change Here
    \link{Code}{https://github.com/zzccppp/GVT.git}
    % \link{Datasets}{https://aaai.org/example/datasets}
    % \link{Extended version}{https://aaai.org/example/extended-version}
\end{links}

\section{Introduction}
\label{sec:introduction}

The \textit{de novo} generation of molecular graphs with desired chemical and physical properties is a cornerstone of modern drug discovery, materials science, and chemical engineering \cite{jin2018junction, faez2021deep}. Molecules are naturally represented as graphs, with atoms as nodes and bonds as edges. These graphs are governed by complex topological features and strict chemical valency rules. Developing generative models that can effectively navigate this intricate chemical space to produce valid, novel, and useful molecules remains a formidable scientific challenge.

Various deep learning paradigms have been applied to this task. While early approaches included Generative Adversarial Networks (GANs) \cite{de2018molgan} and Variational Autoencoders (VAEs) \cite{simonovsky2018graphvae}, score-based diffusion models have recently set a high standard for generation quality \cite{niu2020permutation, jo2022score, yan_swingnn_2024}. However, their iterative sampling process often incurs significant computational overhead and handles the discrete constraints of molecular structures only indirectly. Even hybrid methods aiming to improve this, like PARD \cite{zhao_pard_2024}, which combines diffusion with an autoregressive model, introduces significant framework complexity and still faces challenges with sampling speed. Another prominent approach, Autoregressive (AR) graph generation \cite{you2018graphrnn, jin2018junction}, constructs graphs sequentially. While intuitive, direct AR methods often suffer from error accumulation and struggle to enforce global chemical validity without complex, hard-coded rules.

% 之前的工作，这里总结为Hybrid方法，
Prior work has also explored two-stage generative frameworks that combines discrete latent representations with autoregressive sequence modeling. For instance, DGAE \cite{boget_discrete_2023} investigated a Graph VQ-VAE with an autoregressive component for molecular generation. While promising in their conceptualization, such discrete latent-based approaches have not consistently surpassed the performance of diffusion models, particularly in terms of generation quality and diversity, nor have they fully addressed the limitations of enforcing chemical rules implicitly.

This work introduces Graph VQ-Transformer (GVT), a two-stage framework that leverages the strengths of high-fidelity discrete latent representations and powerful Transformer-based sequence modeling. GVT pushes the boundaries of what two-stage generative models can achieve in molecular design, offering a compelling alternative to computationally intensive continuous diffusion models. Our contributions are:

\begin{itemize}
    \item We design a novel Graph Vector Quantized Variational Autoencoder (VQ-VAE) that achieves exceptionally high reconstruction fidelity on complex molecular datasets. This effectively solves the fidelity bottleneck that limits previous two-stage models.
    \item At the core of our VQ-VAE is a new decoder architecture that synergistically combines Reverse Cuthill-McKee (RCM) for canonical node ordering with a Graph Transformer enhanced by Rotary Positional Encodings (RoPE). This mechanism is crucial, as it allows the decoder to interpret sequential proximity in the latent space as structural information, resolving ambiguities that standard GNNs cannot.
    \item We demonstrate that GVT's high-fidelity latents translate directly to state-of-the-art generative performance. Our full model achieves superior or highly competitive results across major benchmarks, significantly outperforming leading diffusion models on key distribution similarity metrics and presenting a fast, accurate alternative for molecular design.
\end{itemize}

We empirically validate GVT on the QM9\cite{wu2018moleculenet}, ZINC250k\cite{irwin2012zinc}, MOSES\cite{moses_dataset}, and GuacaMol\cite{brown2019guacamol} datasets, showing its effectiveness for both high-fidelity representation learning and efficient, high-quality molecular generation.

\section{Related Work}
\label{sec:related_work}

\subsection{Autoregressive Graph Generation}
Autoregressive (AR) models have been widely adapted for discrete graph structures, typically by constructing a graph step-by-step. This sequential approach inherently acknowledges the discrete nature of graphs but faces a key challenge: graphs lack a single, canonical order for their nodes and edges. The performance of an AR model thus depends heavily on the ordering scheme used to serialize the graph, as it aims to model the probability $p(G) = \sum_{\pi \in \mathcal{P}(G)} p(G, \pi)$ over all possible permutations.

To address this, early AR graph generation models focused on defining a canonical ordering for the nodes before generation. Pioneering works like GraphRNN \cite{you2018graphrnn} and GRAN \cite{liao2019efficient} proposed various strategies, such as using Breadth-First Search (BFS) or sorting nodes by their degree, to create a consistent sequence from a graph structure. These methods aim to transform the complex graph generation task into a more manageable sequential prediction problem.

In a two-stage framework like ours, this principle of ordering is even more critical. Our AR model does not operate on the graph directly, but on a sequence of discrete latent codes produced by the VQ-VAE encoder. Therefore, a consistent and structurally meaningful node ordering is paramount. It ensures that the resulting latent sequence is predictable and that the AR model can effectively learn local dependencies, as structurally adjacent nodes in the graph are mapped to nearby positions in the latent sequence. This stable mapping is the foundation upon which our high-performing generative model is built.

\subsection{Graph Autoencoders and Vector Quantization}
\label{subsec:graph_autoencoders}

Graph autoencoders (GAEs) are crucial for learning low-dimensional latent representations for generation. Early methods used a single, continuous, graph-level vector \cite{simonovsky2018graphvae}, but this approach struggled with complex and inaccurate reconstruction. To simplify this, subsequent work shifted to node-level continuous latent spaces \cite{samanta_nevae_2020, li2018multi}, but this introduced latent-node collapse, where embeddings become indistinguishable and fail to capture local structure.

To solve this, recent efforts have turned to discrete latent spaces using Vector Quantization (VQ), as popularized by VQ-VAEs \cite{oord_neural_2018} and adapted for graph data \cite{yang_vqgraph_2023}. VQ maps continuous vectors to a finite, learned codebook, creating a more structured representation. However, prior Graph VQ-VAE models like DGAE \cite{boget_discrete_2023} were limited by low reconstruction fidelity. A generative model trained on such noisy discrete codes cannot be expected to master complex chemical rules. This analysis motivates our central focus: developing a Graph VQ-VAE capable of exceptionally high-fidelity reconstruction, thereby providing a robust foundation for the subsequent autoregressive model.

\section{Methods}
\label{sec:methods}

\begin{figure*}[t]
    \centering
    \includegraphics[width=0.9\textwidth]{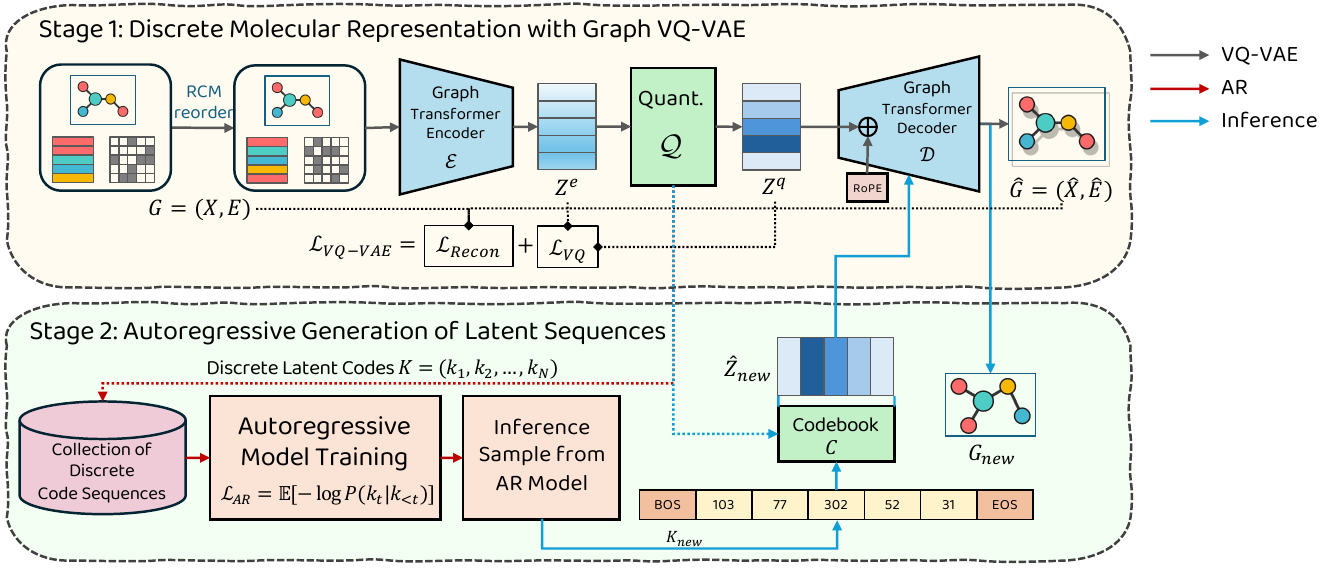}
    \caption{
        An overview of our proposed two-stage framework. 
        \textbf{Stage 1: Graph VQ-VAE.} A molecular graph is first preprocessed with Reverse Cuthill-McKee (RCM) for canonical node ordering. The Graph Transformer-based Encoder maps the graph to continuous latent vectors, which are then quantized into a sequence of discrete codes by the Vector Quantization (VQ) layer. The Decoder, which uniquely uses RoPE to interpret sequential proximity as structural information, reconstructs the graph from these codes. The model is trained end-to-end via a reconstruction and commitment loss.
        \textbf{Stage 2: Autoregressive Generation.} The trained VQ-VAE is used to encode a dataset of molecules into discrete latent sequences. A decoder-only autoregressive Transformer is then trained on these sequences. New molecules are generated by sampling a latent sequence from the AR model and decoding it back into a graph using the pre-trained VQ-VAE decoder.
    }
    \label{fig:framework}
\end{figure*}

Our proposed framework, illustrated in Figure~\ref{fig:framework}, consists of two main stages: (1) learning high-fidelity discrete representations of molecules using a novel Graph VQ-VAE, and (2) training an autoregressive model on these discrete sequences.

\subsection{Stage 1: High-Fidelity Discrete Molecular Representation with Graph VQ-VAE}
\label{subsec:graph_vqvae}

The first stage of our framework is to learn a mapping from a molecular graph $\mathbf{G} = (\mathbf{X} \in \mathbb{R}^{N \times d_x}, \mathbf{E} \in \mathbb{R}^{N \times N \times d_e})$ to a sequence of discrete integer codes $K=(k_1, k_2, \dots, k_N)$, from which $\mathbf{G}$ can be perfectly reconstructed. The success of the entire generative pipeline hinges on the fidelity of this stage. Our design is motivated by solving key challenges inherent in graph representation.

\subsubsection{The Challenge: Structural Ambiguity and Permutation Equivariance}
A core property of Graph Neural Networks (GNNs) is permutation equivariance. While powerful, this property creates a challenge for reconstruction: a standard GNN decoder treats structurally identical nodes 
%(e.g., the two methyl carbons in a propane molecule) 
as indistinguishable. This ambiguity can make it difficult for the decoder to precisely reconstruct specific bond connections, as it lacks a mechanism to differentiate between these symmetric nodes.

\begin{figure}[tb]
    \centering
    \includegraphics[width=1.0\columnwidth]{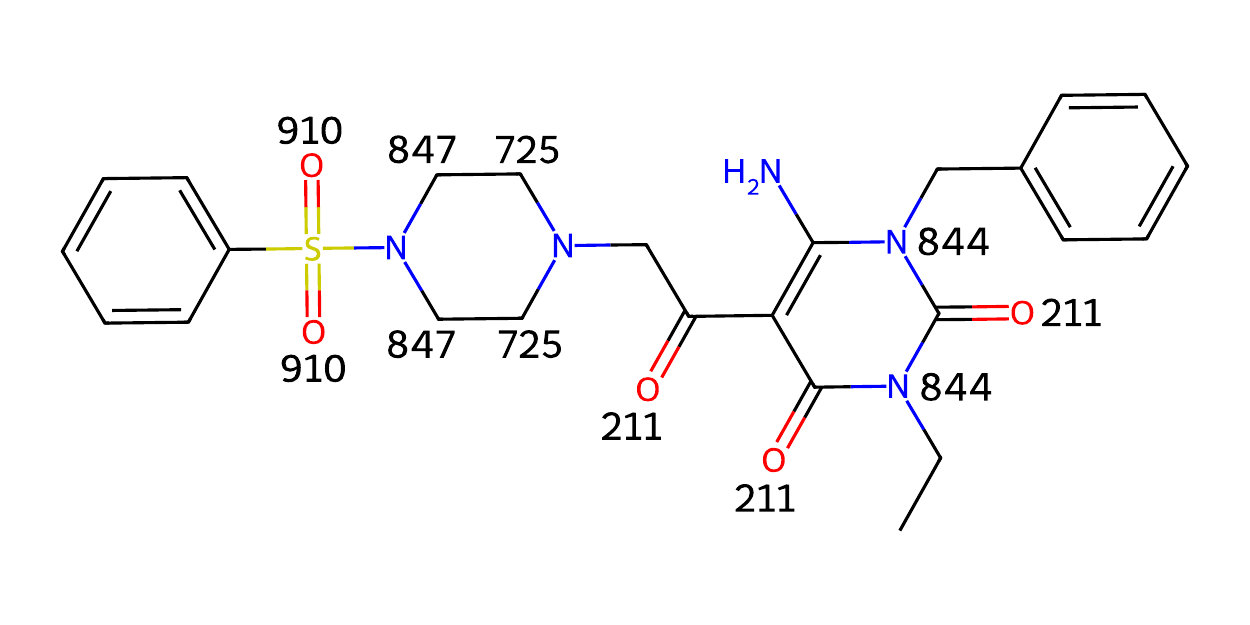}
    \caption{An example of structural ambiguity. Three distinct oxygen atoms are encoded into the same discrete latent code ($k=211$).}
    \label{fig:ambiguity_example}
\end{figure}

To illustrate this challenge with a concrete, empirically observed example, consider the example molecule from the GuacaMol dataset shown in Figure~\ref{fig:ambiguity_example}. After passing through our VQ-VAE encoder, several atoms are mapped to their corresponding discrete codebook indices ($k_i$). Notably, three distinct oxygen atoms, each double-bonded to a carbon atom within a carboxyl group, are all mapped to the identical codebook index, $k=211$.

For a standard, permutation-equivariant GNN decoder, these three nodes are indistinguishable. They possess identical feature information (the latent code for $k=211$) and, within the assumed fully-connected latent graph that the decoder operates on, they hold equivalent topological status. This creates a critical ambiguity: the decoder has no intrinsic information to correctly associate each oxygen with its specific parent carbon atom. It might erroneously attempt to connect one carbon to multiple oxygens, or vice-versa, leading to invalid chemical structures during reconstruction.

% This ambiguity underscores the necessity of breaking the decoder's permutation equivariance and introducing an alternative source of information. By first applying a canonical ordering (RCM), these three nodes, while sharing the same code $k=211$, are assigned unique sequential positions. Subsequently, as we will detail, our RoPE-enhanced decoder is made sensitive to these positions. It can now differentiate between "the node $k=211$ at position 5" and "the node $k=211$ at position 12", for instance. This resolves the ambiguity, drastically reducing the potential for error and enabling the correct, high-fidelity reconstruction of local chemical structures.

\subsubsection{Canonical Node Ordering with Reverse Cuthill-McKee (RCM)}
To resolve this ambiguity, we first establish a consistent, structurally-aware node sequence for every graph. We preprocess all input graphs using the Reverse Cuthill-McKee (RCM) algorithm \cite{cuthill1969reducing}, which reorders nodes to minimize the bandwidth of the adjacency matrix. This canonicalization ensures that structurally proximal nodes are placed close to each other in the sequence, providing a deterministic order that breaks symmetries. Our choice of RCM is empirically validated against other common ordering schemes like Breadth-First Search (BFS) and random ordering, as detailed in the Appendix, where RCM consistently yields the highest reconstruction fidelity. However, this introduces a new challenge: a standard GNN decoder, by its equivariant nature, is designed to ignore this valuable ordering information. Simply feeding it an ordered sequence would be futile.

\subsubsection{Encoder $\mathcal{E}$}
The encoder $\mathcal{E}$ maps an RCM-ordered input graph $\mathbf{G}$ to a set of node-level continuous latent vectors $\mathbf{Z}^e = \{\mathbf{z}_i^e\}_{i=1}^N$, where $\mathbf{z}_i^e \in \mathbb{R}^{d_{c}}$.
Input node features $\mathbf{X}$, edge features $\mathbf{E}$, and Laplacian Positional Encodings $\mathbf{P}$ are processed through $L_{enc}$ layers of a Graph Transformer network \cite{yun2019graph}. The Graph Transformer layers update node and edge representations using attention mechanisms that integrate node, edge, and positional information:
$$ (\mathbf{H}^{(l+1)}, \mathbf{E}^{(l+1)}) = \text{GTLayer}(\mathbf{H}^{(l)}, \mathbf{E}^{(l)}, \mathbf{P}, \mathbf{A}) $$
After the final Graph Transformer layer, we obtain the node representations $\mathbf{H}^{(L_{enc})}$ and edge representations $\mathbf{E}^{(L_{enc})}$. To generate a more comprehensive node embedding, we introduce a fusion step that combines these two sources of information. Specifically, for each node $i$, we first aggregate the features of its incoming edges by computing their mean:
$$ \mathbf{e}_i^{\text{agg}} = \underset{j:(j,i) \in \mathcal{E}}{\text{mean}} \left( \mathbf{e}_{ji}^{(L_{enc})} \right) $$
where $\mathbf{e}_{ji}^{(L_{enc})}$ is the feature vector of the edge from node $j$ to node $i$ produced by the final layer, and $\mathcal{E}$ is the set of edges in the graph. This aggregated edge information is then concatenated with the node's own representation, $\mathbf{h}_i^{(L_{enc})} \in \mathbf{H}^{(L_{enc})}$. The resulting vector is passed through a final linear fusion layer to produce the final node latent vector $\mathbf{z}_i^e$:
$$ \mathbf{z}_i^e = \mathbf{W}_{f} \left( \mathbf{h}_i^{(L_{enc})} \mathbin\Vert \mathbf{e}_i^{\text{agg}} \right) + \mathbf{b}_{f} $$
where $\mathbf{W}_{f}$ and $\mathbf{b}_{f}$ are the trainable weight matrix and bias of the linear fusion layer, and $\mathbin\Vert$ denotes the concatenation operation. The complete set of these vectors $\mathbf{Z}^e = \{\mathbf{z}_i^e\}_{i=1}^N$ constitutes the output of the encoder. This entire encoder architecture, including the fusion stage, is permutation equivariant.

\subsubsection{Vector Quantization Layer $\mathcal{Q}$}
The continuous latent vectors $\mathbf{Z}^e$ are mapped to discrete representations using a standard Vector Quantization (VQ) layer. A learnable codebook $\mathbf{C} = \{\mathbf{c}_k\}_{k=1}^{K_c}$ contains $K_c$ codebook vectors (embeddings) of dimension $d_{c}$. Each $\mathbf{z}_i^e$ is mapped to its closest codebook vector $\mathbf{c}_{k_i}$:
$$ k_i = \arg \min_k ||\mathbf{z}_i^e - \mathbf{c}_k||_2^2 $$
To enable end-to-end training through this non-differentiable operation, the gradient from the quantized vectors $Z_q$ is copied directly to the encoder's output $Z_e$ during the backward pass using a straight-through estimator.
The output is a sequence of quantized vectors $\mathbf{Z}^q = \{\mathbf{z}_i^q\}_{i=1}^N$ where $\mathbf{z}_i^q = \mathbf{c}_{k_i}$, and the corresponding integer indices $K=(k_1, \dots, k_N)$.

\subsubsection{Decoder $\mathcal{D}$: Fusing Order and Structure with RoPE}
Our decoder is specifically designed to bridge the gap between the RCM-imposed node order and the need for structural reconstruction. To achieve this, we employ a Graph Transformer architecture uniquely enhanced with Rotary Position Embeddings (RoPE) \cite{su2024roformer}. This is the core of our VQ-VAE's high fidelity. The synergy is twofold: RCM embeds structural proximity into sequential proximity, and RoPE allows the attention mechanism to directly process this sequential information as a relative positional signal.

The reconstruction process proceeds in two steps:
\begin{enumerate}
    \item \textbf{Injecting Relative Positional Awareness via RoPE:} We first transform the raw sequence of quantized vectors $\mathbf{Z}^q$ into a position-aware sequence $\mathbf{Z}^{q'}$. RoPE achieves this by rotating each vector $\mathbf{z}_i^q$ based on its absolute position $i$. The critical property of RoPE is that the subsequent attention score between any two vectors at positions $i$ and $j$ becomes an elegant function of their content and their relative distance, $i-j$. Since RCM ordering ensures that a small relative distance implies structural closeness, RoPE provides the decoder with a direct and powerful signal for graph topology. This position-aware sequence $\mathbf{Z}^{q'}$ serves as the initial node representations $\mathbf{H}^{(0)}$ for the decoder.

    \item \textbf{Iterative Graph Reconstruction:} With position-aware node embeddings $\mathbf{H}^{(0)}$, the Graph Transformer decoder iteratively refines the graph structure. First, initial edge features $\mathbf{E}^{(0)}$ are predicted by a shared MLP that processes pairs of node embeddings, ensuring symmetry. These initial predictions, along with $\mathbf{H}^{(0)}$, are fed into a stack of $L_{dec}$ Graph Transformer layers. Each layer updates both node and edge representations: $(\mathbf{H}^{(l+1)}, \mathbf{E}^{(l+1)}) = \text{GTLayer}(\mathbf{H}^{(l)}, \mathbf{E}^{(l)})$. After the final layer, the node representations $\mathbf{H}^{(L_{dec})}$ become the reconstructed node features $\mathbf{\hat{X}}$, and the final edge representations $\mathbf{E}^{(L_{dec})}$ are symmetrized to form the reconstructed adjacency and edge-type matrix $\mathbf{\hat{A}}$.
\end{enumerate}
This design allows the decoder to leverage the canonical ordering effectively, leading to highly accurate graph reconstruction.

\subsubsection{VQ-VAE Training}
The VQ-VAE is trained end-to-end by minimizing a combined loss function, which includes reconstruction terms for node and edge features, as well as the standard VQ commitment loss:
\begin{equation}
    \begin{aligned}
        \mathcal{L}_{\text{VQ-VAE}} &= \lambda_{\text{node}} \mathcal{L}_{\text{node}} + \lambda_{\text{edge}} \mathcal{L}_{\text{edge}}
        \\ &+ ||\text{sg}(\mathbf{Z}^e) - \mathbf{Z}^q||_2^2 + \beta ||\mathbf{Z}^e - \text{sg}(\mathbf{Z}^q)||_2^2
    \end{aligned}
\end{equation}
Here, $\mathcal{L}_{\text{node}}$ and $\mathcal{L}_{\text{edge}}$ are cross-entropy losses for reconstruction. The latter two terms align the encoder output with the codebook, where $\text{sg}(\cdot)$ is the stop-gradient operator. A detailed summary of the two-stage training procedure is provided as pseudocode in the Appendix.

\subsection{Stage 2: Autoregressive Generation of Latent Sequences}
\label{subsec:autoregressive_models}

Once the Graph VQ-VAE is trained, its encoder $\mathcal{E}$ and quantizer $\mathcal{Q}$ are used to convert a dataset of RCM-ordered molecular graphs into their corresponding sequences of discrete latent codes $K = (k_1, k_2, \dots, k_N)$.

We then train a Transformer-based decoder-only autoregressive model (similar to GPT2 \cite{radford2019language}) to model the distribution $P(K)$. The model predicts the next code $k_t$ given the previous codes $k_{<t}$:
$$ P(K) = \prod_{t=1}^{N} P(k_t | k_{<t}; \theta_{AR}) $$
The AR model is trained by minimizing the negative log-likelihood (cross-entropy loss) of the true sequences:
$$ \mathcal{L}_{\text{AR}} = -\sum_{\text{seq } K} \sum_{t=1}^{N} \log P(k_t | k_{<t}; \theta_{AR}) $$
The RCM ordering provides a canonical and structurally relevant sequence for the AR model, simplifying the learning of dependencies, especially local connectivity patterns which are often reflected in nearby tokens.

\section{Experiments}

\subsection{The Critical Role of RoPE in Reconstruction}
% We first assess the expressive power of our Graph VQ-VAE through its perfect graph reconstruction capability. We use a strict 0-error rate, defined as the percentage of test graphs flawlessly rebuilt in both node features and adjacency.

% To quantify the impact of Rotary Position Embeddings (RoPE), we conducted a targeted ablation study. 
% The results in Figure \ref{fig:reconstruction_ablation} reveal a crucial trend: the performance gain from RoPE is strongly correlated with dataset complexity.

% On the relatively simple QM9 dataset, our model without RoPE already achieves a high score (99.56\%), and the inclusion of RoPE offers a modest but valuable improvement. However, this performance gap widens dramatically on larger and more structurally diverse datasets. For instance, on ZINC250k, RoPE boosts the success rate by over 22 absolute percentage points. The effect is most pronounced on GuacaMol, where the reconstruction rate catapults from a failing 51.38\% to a near-perfect 99.84%.

% This evidence strongly suggests that while our base Transformer architecture is effective, RoPE is the critical component that enables the decoder to robustly interpret the sequential latent codes for complex graph topologies. The near-perfect final scores across all benchmarks validate that our full model produces high-fidelity discrete representations suitable for challenging downstream generative tasks.

\begin{figure}[tb]
    \centering
    \includegraphics[width=1.0\columnwidth]{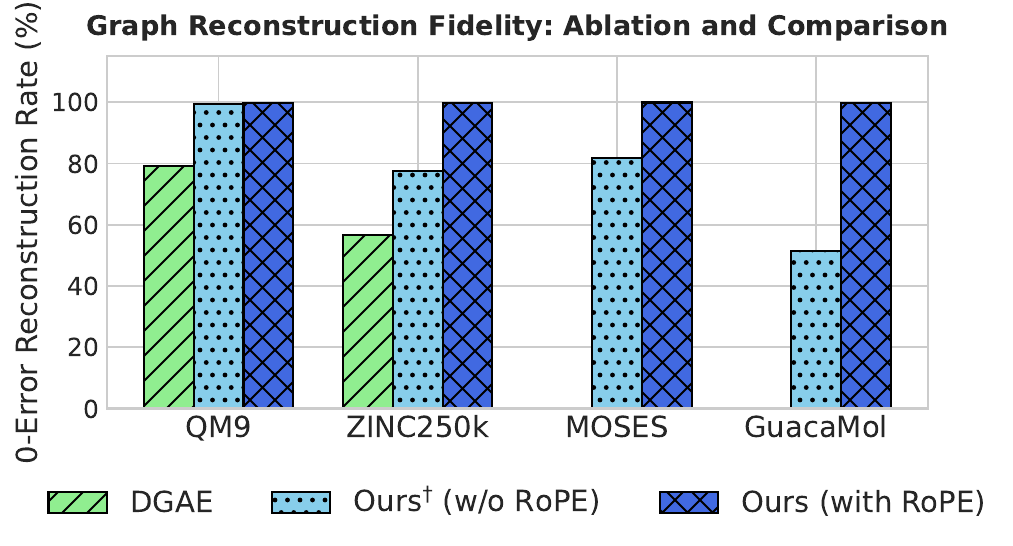}
    \caption{0-Error Reconstruction Rate (\%) on test sets. Our full model (GVT with RoPE) achieves near-perfect reconstruction, drastically outperforming both the previous DGAE's GAE and our own architecture without the crucial RoPE component, especially on complex datasets.}
    \label{fig:reconstruction_ablation}
\end{figure}

Our central hypothesis is that near-perfect reconstruction is the cornerstone of a successful two-stage generative model. A VQ-VAE that produces flawed latents cannot effectively teach a subsequent autoregressive model the complex rules of chemistry. To validate this and dissect our architectural contributions, we measure the 0-error reconstruction rate—the percentage of flawlessly rebuilt graphs—and present our findings in Figure~\ref{fig:reconstruction_ablation}.

% DGAE QM9:0.7926, ZINC250k:0.5678

Our results confirm this hypothesis in a step-by-step manner. First, the DGAE model validates the limitations of prior work, correctly reconstructing only 79.26\% of QM9 and 56.78\% of ZINC250k molecules. Our GVT architecture without RoPE already shows significant improvement (99.56\% on QM9), but struggles with complex datasets like GuacaMol (51.38\%), indicating that RCM ordering alone is insufficient. The critical role of RoPE is undeniable: activating it pushes the reconstruction rate to near-perfection across all benchmarks, including 99.89\% on ZINC250k and 99.84\% on GuacaMol. This dramatic leap, especially the over 48-point gain on GuacaMol, provides conclusive evidence that our synergistic design effectively solves the fidelity problem.

In summary, this analysis validates that our combination of a Graph Transformer, RCM ordering, and particularly RoPE, creates the high-fidelity discrete representations necessary for robust molecular generation. This solid foundation is what enables the success of our subsequent autoregressive model. In the Appendix, we further explore the impact of another key design choice, the codebook size, demonstrating its effect on reconstruction performance and training convergence.

\subsection{Molecule Graph Generation}

\begin{table*}[ht]
\centering
\begin{tabular}{lccccccccc}
\toprule
\multirow{2}{*}{Methods} & \multirow{2}{*}{Class} & \multicolumn{4}{c}{QM9 } & \multicolumn{4}{c}{ZINC250k} \\
\cmidrule(lr){3-6} \cmidrule(lr){7-10}
 & & Val. $\uparrow$ & Uni. $\uparrow$ & FCD $\downarrow$ & NSPDK $\downarrow$ & Val. $\uparrow$ & Uni. $\uparrow$ & FCD $\downarrow$ & NSPDK $\downarrow$ \\
\midrule
GraphAF & \multirow{2}{*}{Flow-Based} & 57.16 & 83.78 & 5.384 & 2.10e-2 & 68.47 & 99.01 & 16.023 & 4.40e-2 \\
GraphDF & & 79.33 & 95.73 & 11.283 & 7.50e-2 & 41.84 & 93.75 & 40.51 & 3.54e-1 \\
\midrule
GDSS & \multirow{4}{*}{Diffusion} & 90.36 & 94.70 & 2.923 & 4.40e-3 & 97.35 & 99.76 & 11.398 & 1.80e-2 \\
DiGress & & 95.43 & 93.78 & 0.643 & 7.28e-4 & 84.94 & 99.21 & 4.88 & 8.75e-3 \\
SwinGNN & & 99.68 & 95.92 & \textbf{0.169} & 4.02e-4 & 87.74 & 99.98 & 5.219 & 7.52e-3 \\
GLAD & & 97.12 & 97.52 & 0.201 & \textbf{3e-4} & 81.81 & \textbf{100} & 2.54 & \textbf{2.00e-3}  \\ 
\midrule
PARD & \multirow{2}{*}{Hybrid} & 97.5 & 95.8 & - & - & 95.23 & 99.99 & 1.98 & - \\
DGAE & & 92.0 & \textbf{97.61} & 0.86 & 1.50e-3 & 77.9 & 99.94 & 4.4 & 7.00e-3 \\
\midrule
Ours & VQVAE+AR & \textbf{99.76} & {95.94} & 0.87 & 6.0e-4 & \textbf{99.57} & 97.55 & \textbf{1.16} & 6.89e-2 \\
\bottomrule
\end{tabular}
\caption{QM9 and ZINC250k Generation Results.}
\label{tab:combined_results}
\end{table*}

\begin{table*}[ht]
\centering
\begin{tabular}{lllllllll}
\toprule
Model & Class & Val. $\uparrow$ & Unique $\uparrow$ & Novel $\uparrow$ & Filters $\uparrow$ & FCD $\downarrow$ & SNN $\uparrow$ & Scaf $\uparrow$ \\
\midrule
VAE & SMILES & 97.7 & 99.8 & 69.5 & \textbf{99.7} & \underline{0.57} & \underline{0.58} & 5.9 \\
JT-VAE & Fragment & \textbf{100} & 100 & \textbf{99.9} & 97.8 & 1.00 & 0.53 & 10 \\
GraphINVENT & AR & 96.4 & 99.8 & - & 95.0 & 1.22 & 0.54 & 12.7 \\
\midrule
ConGress & Diffusion & 83.4 & 99.9 & \underline{96.4} & 94.8 & 1.48 & 0.50 & \textbf{16.4} \\
DiGress & Diffusion & 85.7 & 100 & 95.0 & 97.1 & 1.19 & 0.52 & \underline{14.8} \\
PARD & AR+Diffusion & 86.8 & 100 & 78.2 & 99.0 & 1.00 & 0.56 & 2.2 \\
\midrule
Ours & VQVAE+AR & \underline{99.42} & \textbf{100} & 61.43 & \underline{99.59} & \textbf{0.16} & \textbf{0.64} & 3.4 \\
\bottomrule
\end{tabular}
\caption{Molecule generation on MOSES. All metrics $\uparrow$ except FCD $\downarrow$.}
\label{tab:MOSES_Result}
\end{table*}

% 搞清楚NAGVAE和MCTS的方法类型
\begin{table*}[ht]
\centering
\begin{tabular}{lllllllll}
\toprule
Model & Class & Valid $\uparrow$ & Unique $\uparrow$ & Novel $\uparrow$ & KL div $\uparrow$ & FCD $\uparrow$ \\
\midrule
LSTM & SMILES & 95.9 & \textbf{100} & 91.2 & \underline{99.1} & \textbf{91.3} \\
NAGVAE & Autoregressive & 92.9 & 95.5 & \textbf{100} & 38.4 & 0.9 \\
MCTS & Search-based & \textbf{100} & \textbf{100} & 95.4 & 82.2 & 1.5 \\
\midrule
DiGress & Diffusion & 85.2 & \textbf{100} & \underline{99.9} & 92.9 & 68.0 \\
\midrule
Ours & VQVAE+AR & \underline{97.46} & \underline{99.9} & 82.39 & \textbf{99.61} & \underline{85.50} \\
\bottomrule
\end{tabular}
\caption{Molecule generation on GuacaMol. We report scores from benchmark, so that higher is better for all metrics.}
\label{tab:Guacamol_Result}
\end{table*}

\textbf{Datasets.} We experiment with four different molecular datasets: QM9\cite{wu2018moleculenet}, ZINC250k\cite{irwin2012zinc}, MOSES\cite{moses_dataset}, GuacaMol\cite{brown2019guacamol}.

For experimental setup, we use a fixed 80\%-20\% training-test split for the QM9 and ZINC250k datasets. For MOSES and GuacaMol, we adhere to their officially defined data splits to ensure fair comparison with previous work.
Following \cite{jo2022score}, we remove hydrogen atoms and kekulize molecules by RDKit \cite{landrum2016rdkit}. Detailed hyperparameters for our models, including layer counts, dimensions, and training configurations for each dataset, are provided in the Appendix.

\textbf{Baselines.}
We benchmark our model against a diverse set of baselines from state-of-the-art graph generative model families. The specific models for comparison vary depending on the dataset and its standard evaluation protocol.

\noindent\textbf{For QM9 and ZINC250k,} we compare against a comprehensive suite of recent graph-native generative models. These are grouped into three main categories:
\begin{itemize}
    \item \textbf{Flow-based Models:} This category includes generative flow models such as GraphAF \cite{shi2020graphaf} and GraphDF \cite{luo2021graphdf}.
    
    \item \textbf{Diffusion Models:} We select a range of representative diffusion-based approaches, namely the score-based GDSS \cite{jo2022score}, the discrete DiGress \cite{vignac_digress_2022}, SwinGNN \cite{yan_swingnn_2024}, and GLAD \cite{nguyen_glad_2025}.
    
    \item \textbf{Hybrid Autoregressive Models:} In this category, we include DGAE \cite{boget_discrete_2023}, which is architecturally the most analogous to our approach as it also combines a VQ-VAE with an autoregressive (AR) framework. We also compare against PARD \cite{zhao_pard_2024}, which represents a different hybrid strategy that integrates AR modeling with a diffusion backbone. 
\end{itemize}

\noindent\textbf{For the MOSES benchmark,} in addition to recent one-shot methods like DiGress and PARD, we follow the standard protocol by also comparing against a wider array of established models. As shown in Table~\ref{tab:MOSES_Result}, this includes methods that operate on different molecular representations, such as SMILES-based VAEs, fragment-based models like JT-VAE, and the classic autoregressive model GraphINVENT.

\noindent\textbf{For the GuacaMol benchmark,} we compare against several models from its standard leaderboard to assess performance on goal-directed tasks. As shown in Table~\ref{tab:Guacamol_Result}, these baselines include the SMILES-based LSTM, as well as one-shot methods like NAGVAE and the search-based MCTS.

\textbf{Metrics.}
To comprehensively evaluate the performance of our model against baselines, we employ a set of widely-recognized metrics for molecular generation, sourced from standard benchmarks like GuacaMol \cite{brown2019guacamol} and MOSES \cite{moses_dataset}. Core metrics include \textbf{Validity (Val.)}, the percentage of chemically valid molecules; \textbf{Uniqueness (Uni.)}, the proportion of unique structures among valid ones to detect mode collapse; and \textbf{Novelty}, the percentage of valid unique molecules not present in the training set. As explained in \cite{vignac_digress_2022}, Novelty is not reported in main table. To assess distribution similarity, we use the \textbf{Fréchet ChemNet Distance (FCD)}, where a lower score indicates higher similarity. For a deeper structural comparison, we report the \textbf{Network-based Shortest Path Difference Kernel (NSPDK)}, where lower values are also better. The MOSES benchmark provides a more holistic suite, including \textbf{Scaffold Similarity (SNN)} and \textbf{Scaffold Uniqueness (Scaf)} for scaffold-level analysis, and a \textbf{Filters} metric which reports the percentage of molecules passing medicinal chemistry screens. The GuacaMol benchmark introduces a goal-directed \textbf{KL Divergence (KL div)} score, which measures the similarity of physicochemical property distributions, with higher scores indicating a better match.

\subsection{Main Results}
We present the generation results on QM9, ZINC250k, MOSES, and GuacaMol in Table~\ref{tab:combined_results},~\ref{tab:MOSES_Result}, and~\ref{tab:Guacamol_Result}, respectively. For fair comparison, all baseline results reported in the tables are sourced from their original publications. The results demonstrate that our autoregressive model achieves highly competitive, and often state-of-the-art, performance, particularly on large and challenging benchmarks. Our approach consistently matches or exceeds the performance of leading diffusion models, traditional rule-based autoregressive methods, and SMILES-string-based models.

\noindent\textbf{Performance on QM9 and ZINC250k.}
On the small QM9 dataset, our model achieves the highest validity (99.76\%) among all baselines. More importantly, on the large-scale and more challenging ZINC250k benchmark, our model demonstrates its superiority in distribution learning. It achieves a state-of-the-art FCD score of 1.16, significantly outperforming all diffusion-based counterparts such as PARD (1.98) and GLAD (2.54). Furthermore, it attains the highest validity score of 99.38\%, confirming its ability to learn robust chemical rules for complex molecules.

% 需要对模型重新分类
\noindent\textbf{Performance on MOSES Benchmark.}
This strong performance continues on the MOSES benchmark, another large-scale dataset. As shown in Table~\ref{tab:MOSES_Result}, our model sets a new state-of-the-art in distribution similarity with an FCD score of 0.16, which is over 3 times better than the next best competitor (VAE, 0.57). It also achieves the highest diversity score (SNN of 0.64) and passes nearly all medicinal chemistry filters (99.59\%). This result is particularly noteworthy as it surpasses not only other diffusion graph generation methods (e.g., DiGress), but also models that operate on different representations like SMILES strings (VAE) and fragments (JT-VAE).

\noindent\textbf{Performance on GuacaMol Benchmark.}
On the GuacaMol distribution learning benchmark, GVT demonstrates state-of-the-art performance. As shown in Table~\ref{tab:Guacamol_Result}, it achieves the highest KL Divergence score (99.61), significantly outperforming other graph-native models like NAGVAE and MCTS that fail on this task despite high validity. Crucially, GVT also maintains a high FCD score (85.50), showcasing a holistic ability to replicate both the statistical (physicochemical) and structural properties of the target distribution, a feat not guaranteed by simpler 1D SMILES-based models.

\noindent\textbf{Discussion on Model Limitations}
While GVT shows robust performance, we acknowledge two primary limitations inherent to its design, which highlight areas for future improvement.

First, on ZINC250k, GVT's excellent FCD score contrasts with a poorer NSPDK score compared to diffusion models. This suggests high fidelity in local chemical features but a mismatch in global graph topology. We attribute this to the model's inherently local focus. The autoregressive generation of a token sequence, even one ordered by RCM, excels at predicting local connectivity but may not adequately enforce the global topological constraints measured by NSPDK—a task where holistic update models like diffusion may have an advantage.

Second, the lower Novelty scores on MOSES and GuacaMol reflect a deliberate design trade-off prioritizing distribution fidelity over generative exploration. GVT's training objective, combining near-perfect reconstruction with maximum likelihood, incentivizes the model to master the training distribution. While this leads to state-of-the-art distribution-similarity scores (FCD/KL-Div), it naturally reduces the likelihood of generating highly novel scaffolds from the low-probability tails of the learned distribution. However, this limitation is partially offset by GVT's high sampling efficiency. Its fast generation speed allows for extensive over-sampling, where generating a vast number of candidates in a short time significantly increases the chance of discovering unique and novel structures, a practical remedy not always feasible for slower models. Furthermore, as demonstrated in the Appendix, the novelty can be substantially improved by adjusting sampling parameters like temperature and top-k, which encourages the model to explore less probable regions of the latent space without a significant drop in validity.

In summary, across multiple challenging benchmarks, our autoregressive model proves to be a powerful and effective generative method. It not only overcomes the limitations often associated with traditional AR models but also establishes a new level of performance that is highly competitive with the current state-of-the-art diffusion models.

\subsection{Generation Time}

To evaluate the generation efficiency of our model against existing methods, we conducted a speed benchmark. For each model, we generated 10,000 molecules using the respective official codebases and models pre-trained on the QM9 dataset. All experiments were performed on a single NVIDIA RTX 4090 GPU to ensure a fair comparison.

\begin{figure}[tb]
    \centering
    \includegraphics[width=\columnwidth]{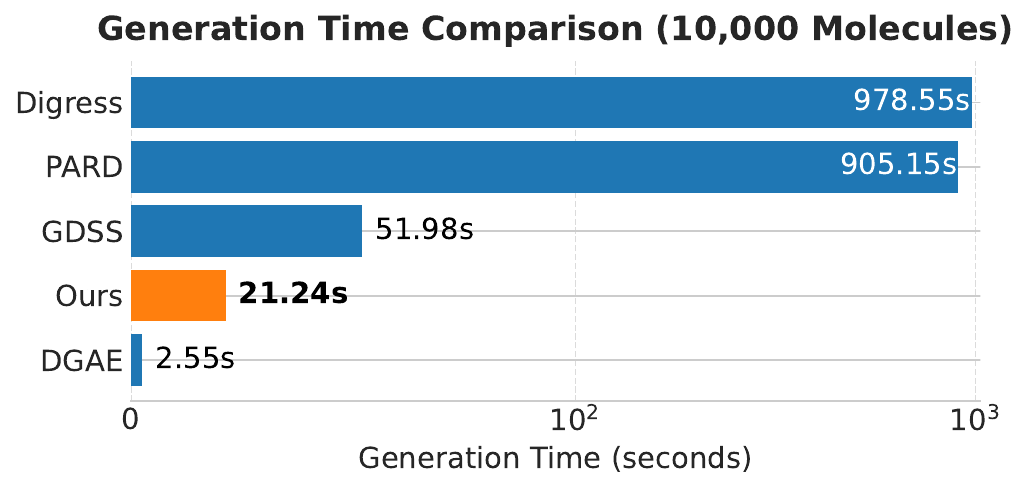}
    \caption{Comparison of generation time for sampling 10,000 molecules on the QM9 dataset. All models were benchmarked on an NVIDIA RTX 4090 GPU. The x-axis is on a logarithmic scale to better visualize the wide range of speeds. Our model shows a competitive generation time, significantly outperforming other diffusion-based methods.}
    \label{fig:gentime}
\end{figure}

The total time taken to generate the full set of molecules is recorded for each model. As shown in Figure~\ref{fig:gentime}, our model demonstrates a highly competitive generation speed. It is significantly faster than diffusion-based models like Digress (978.55s), GDSS (51.98s), and Autoregressive+Diffusion model like PARD(905.15s). While the autoregressive model DGAE is the fastest (2.55s), our approach (21.24s) achieves a practical and efficient generation time, striking a strong balance between speed and the quality afforded by our two-stage framework.

% Sample 10000 Molecules by model trained on QM9 Dataset. Comparision the generation time. Official Code, RTX4090, 

% Digress: 978.55s 
% GDSS: 51.98s
% PARD: 905.15s
% DGAE: 2.5467
% Ours: 21.24s

% 效果实验/速度实验/消融实验

\section{Conclusion}

In this paper, we introduced the Graph VQ-Transformer (GVT), a two-stage framework for fast and high-quality molecular generation. Our core contribution is a novel Graph VQ-VAE that leverages canonical node ordering via RCM and a RoPE-enhanced decoder to achieve exceptionally high-fidelity discretization of molecular structures. By training an autoregressive Transformer on these superior discrete latents, GVT achieves near-perfect graph reconstruction and sets a new state-of-the-art on major benchmarks, including ZINC250k, MOSES, and GuacaMol. Its outperformance on key distribution metrics, such as FCD and KL Divergence, establishes that a well-designed discrete latent variable model can match and even exceed the performance of computationally intensive diffusion models, offering a powerful and efficient pathway for generative chemistry.

\section{Acknowledgments}

This work was supported by the National Natural Science
Foundation of China [62071154].

\bibliography{aaai2026}

\end{document}